\documentclass[10pt,twocolumn,letterpaper]{article}

\usepackage[pagenumbers]{iccv} 
\usepackage{colortbl} 
\usepackage{xcolor}    
\usepackage{pifont}
\usepackage{ifsym}
\usepackage{fontawesome5}
\usepackage{fancyhdr}
\usepackage{marvosym}
\usepackage{fontawesome5}
\definecolor{iccvblue}{rgb}{0.21,0.49,0.74}
\usepackage[pagebackref,breaklinks,colorlinks,allcolors=iccvblue]{hyperref}

\def\paperID{*} 

\renewcommand{\ie}{\emph{i.e.}}
\usepackage{multirow}  

\title{Audio-visual Controlled Video Diffusion with Masked Selective State Spaces Modeling for Natural Talking Head Generation}
\author{Fa-Ting Hong$^{1,2}$ \quad Zunnan Xu$^{2,3}$ \quad Zixiang Zhou$^2$\quad Jun Zhou$^{2}$ \\ Xiu Li$^3$ \quad Qin Lin$^2$\quad Qinglin Lu$^2$\quad Dan Xu$^{1,}${\textsuperscript{\Letter}}\\
\vspace{-12pt}
\and
$^1$HKUST \quad $^2$Tencent \quad $^3$Tsinghua University\\ \\
\vspace{-10pt}
\href{https://harlanhong.github.io/publications/actalker/index.html}{{\includegraphics[height=0.4cm]{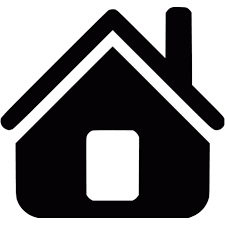}}}
\quad 
\href{https://arxiv.org/abs/2504.02542}{{\includegraphics[height=0.4cm]{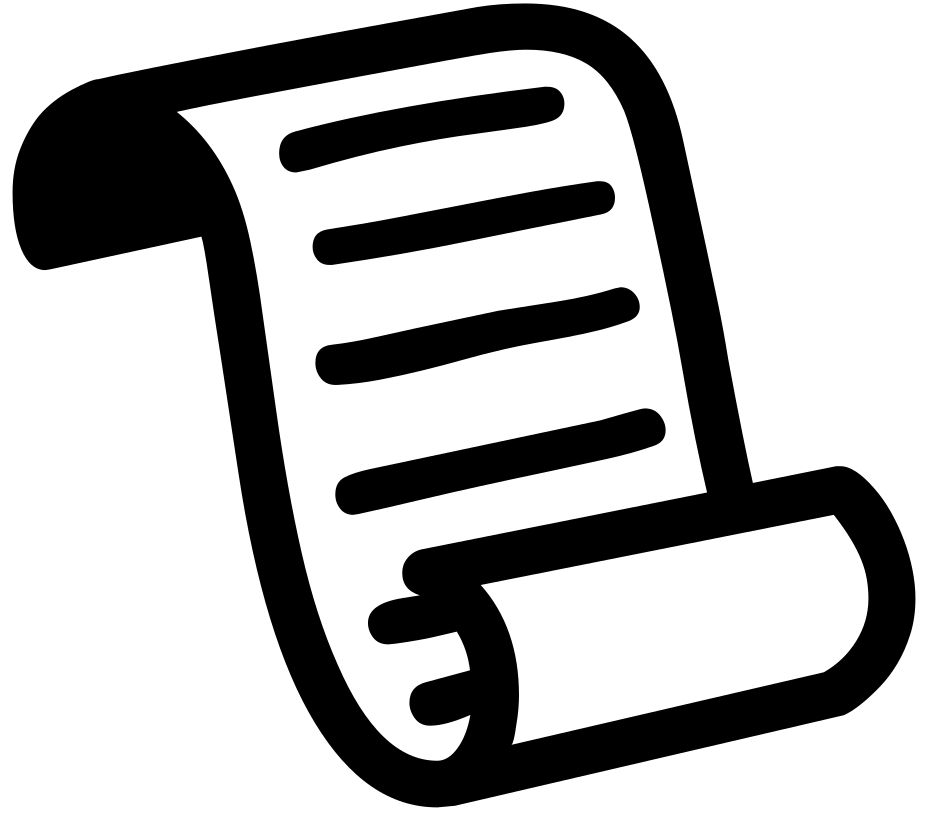}}}
\quad 
\href{https://github.com/harlanhong/ACTalker}{{\includegraphics[height=0.4cm]{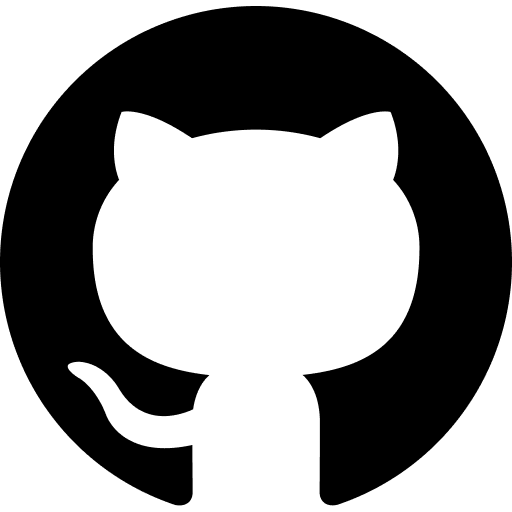}}}
\quad
\href{https://huggingface.co/papers/2504.02542}{{\includegraphics[height=0.4cm]{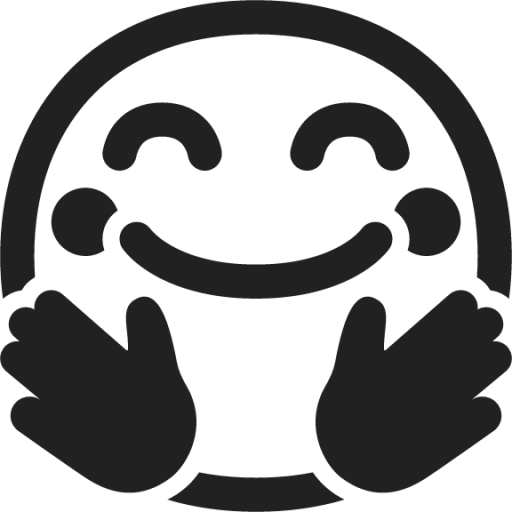}}}
}
\begin{document}
\twocolumn[{
\renewcommand\twocolumn[1][]{#1}
\renewcommand{\thefootnote}{\fnsymbol{footnote}}
\maketitle
\begin{center}
    \centering
    \captionsetup{type=figure}
    \vspace{-15pt}
    \includegraphics[width=0.98\textwidth]{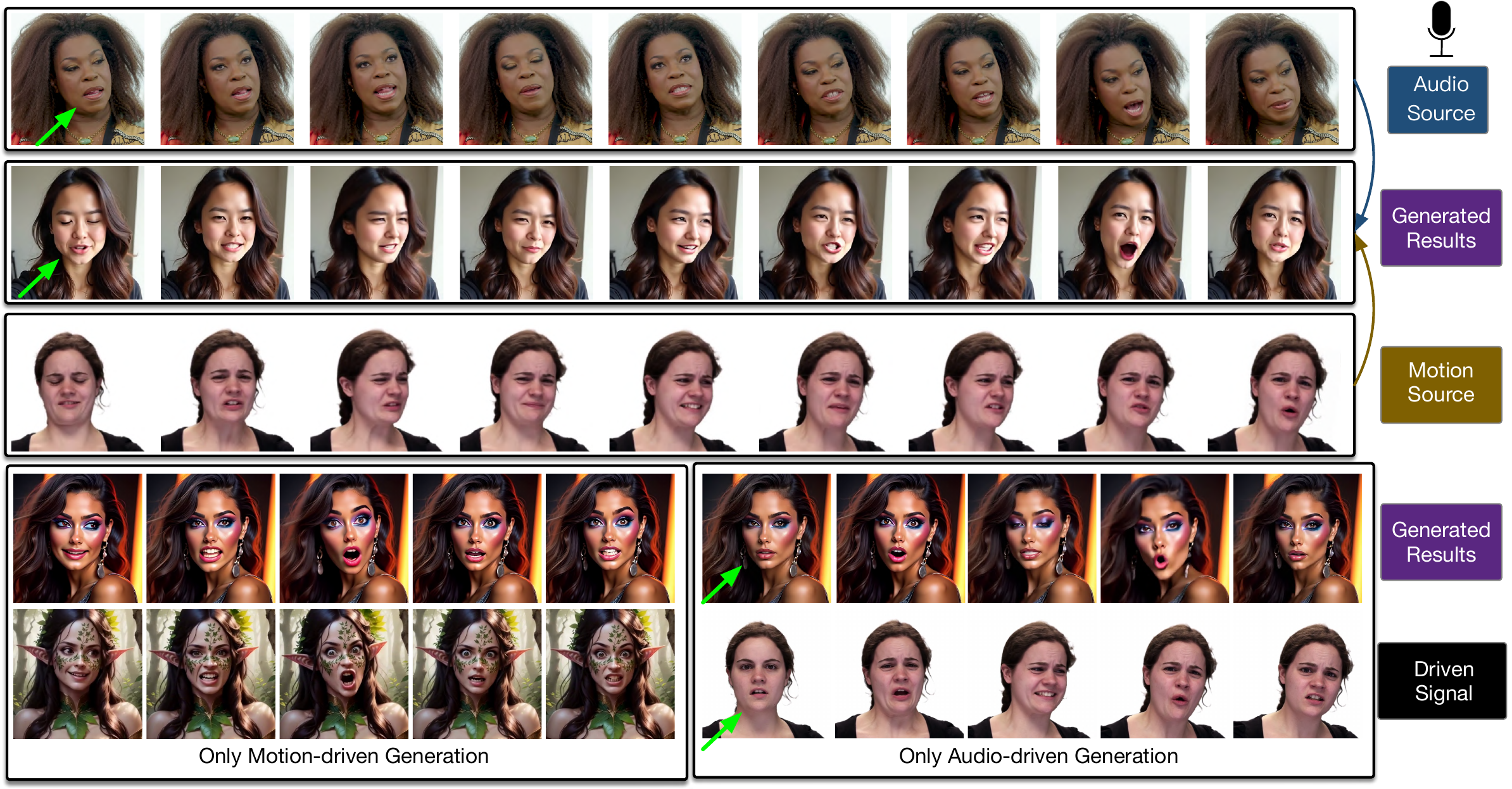}
    \vspace{-8pt}
    \captionof{figure}{In this work, we aim to develop a framework that not only generates videos driven by multiple signals without causing control conflicts in the facial region (first three rows) but also supports video generation driven by a single signal (last two rows).
    }
    \label{fig:teaser}
\end{center}
}]

\begin{abstract}
Talking head synthesis is vital for virtual avatars and human-computer interaction. However, most existing methods are typically limited to accepting control from a single primary modality, restricting their practical utility. To this end, we introduce \textbf{ACTalker}, an end-to-end video diffusion framework that supports both multi-signals control and single-signal control for talking head video generation. For multiple control, we design a parallel mamba structure with multiple branches, each utilizing a separate driving signal to control specific facial regions. A gate mechanism is applied across all branches, providing flexible control over video generation. To ensure natural coordination of the controlled video both temporally and spatially, we employ the mamba structure, which enables driving signals to manipulate feature tokens across both dimensions in each branch. Additionally, we introduce a mask-drop strategy that allows each driving signal to independently control its corresponding facial region within the mamba structure, preventing control conflicts. Experimental results demonstrate that our method produces natural-looking facial videos driven by diverse signals and that the mamba layer seamlessly integrates multiple driving modalities without conflict. The project website can be found at \href{https://harlanhong.github.io/publications/actalker/index.html}{HERE}.
\end{abstract}    
\vspace{-5pt}
\section{Introduction}
\vspace{-5pt}
\label{sec:intro}
Talking head generation~\cite{chung2017lip, siarohin2019first, hong2022depth, hong2023implicit, liu2023moda,Synergizingcvpr25,hong2023dagan++, zhou2021pose, zhou2025fireedit, xu2025hunyuanportrait,lin2025mvportrait} aims to create realistic portrait videos driven by specific input signals. Audio and facial motion are the two primary driving signals for the talking head generation task. In this work, we aim to develop a framework capable of generating portrait videos with either single signal control or simultaneous control of both signals.


Most existing methods typically use a single primary signal to control video generation. They either use audio to control lip movements~\cite{prajwal2020lip, tian2025emo, jiang2024loopy, guan2023stylesync, zhou2021pose, ji2021audio}, or rely on facial motion to govern overall facial dynamics~\cite{hong2023dagan++, hong2022depth, hong2023implicit, siarohin2019first}. 
Furthermore, some studies~\cite{drobyshev2024emoportraits, yin2022styleheat} have focused on developing unified frameworks that support various control signals for video generation. However, they still allow only one signal to drive the generation at a time during inference.

Therefore, generating a portrait video driven by both audio and facial motion remains a significant challenge. Two critical issues must be addressed for effective multi-control: 1) Control conflicts. Audio signals usually have a strong influence on the mouth region and slightly affect the expression of the face, while the facial motion signals can accurately control the facial expression. When both signals are applied simultaneously without resolving their conflicts, the resulting facial expression tends to favor the strongest one. And when signals are applied in a sequential manner, the model may prioritize the more recent one, especially if the signals are in conflict or affect overlapping facial areas. Solving control conflicts is difficult because it requires balancing and blending these two distinct types of signals—one that controls the lower face (mouth) and the other that governs the entire facial expression—without allowing one to dominate the other. 
2) Control signals aggregation. 
Current video diffusion models~\cite{jiang2024loopy, xu2024hallo,tang2025human} typically use attention modules~\cite{vaswani2017attention} to integrate control signals with intermediate features along the temporal and spatial dimensions separately. This separate processing can miss the interactions between temporal and spatial dimensions, leading to less coherent transitions and spatial inconsistencies. Moreover, when control signals are integrated with flattened spatio-temporal features, the attention map becomes extremely large due to the high number of tokens,  especially for longer videos. Thus, finding an efficient way to combine these signals both temporally and spatially remains a critical challenge.



To address the challenges outlined above, we propose the Audio-visual Controlled Video Diffusion model, coined as ACTalker, an end-to-end framework that integrates spatial-temporal features with multiple control signals for photorealistic and expressive talking head generation. To enable the control signals to interact with intermediate video features in both the temporal and spatial dimensions simultaneously, we introduce a selective state-space model (SSM) to aggregate the flattened temporal-spatial feature tokens with the control signals, providing a more computationally efficient alternative to the attention mechanism~\cite{vaswani2017attention}. Furthermore, to facilitate learning, we employ a mask-drop strategy that discards irrelevant feature tokens outside the control regions, enhancing the effectiveness of the driving signals and improving the generated content within the control regions. Importantly, each driving signal is responsible only for the facial regions indicated by a manually specified mask, addressing the control conflict issues among audio and visual control. The SSM structure and mask-drop strategy together form the Mask-SSM unit in our framework, which handles facial control with a single signal in specific regions.
To enable control by multiple signals, we design a parallel-control mamba layer (PCM) consisting of multiple parallel Mask-SSMs. The PCM layer aggregates intermediate features with each driving signal in a spatio-temporal manner within a single branch. To allow simultaneous control by both audio and facial motion while maintaining the flexibility to control by a single signal when needed, we introduce a gating mechanism in each branch, which is randomly set to open during training. This provides flexible control over the generated video, as the gate can be opened or closed during inference, enabling manipulation based on any chosen signals.

We conduct extensive experiments and ablation studies to validate the effectiveness of our proposed method. The experimental results show that our approach not only outperforms existing methods in single-signals control talking head video generation, but also resolves the condition conflict problem to achieve multiple signals control. Our ablation studies demonstrate that our designed mamba structure effectively integrates multiple driving signals with different feature tokens across distinct facial regions, enabling fine-grained signals control without conflicts.
\begin{figure*}[t]
  \centering
    \includegraphics[width=0.98\linewidth]{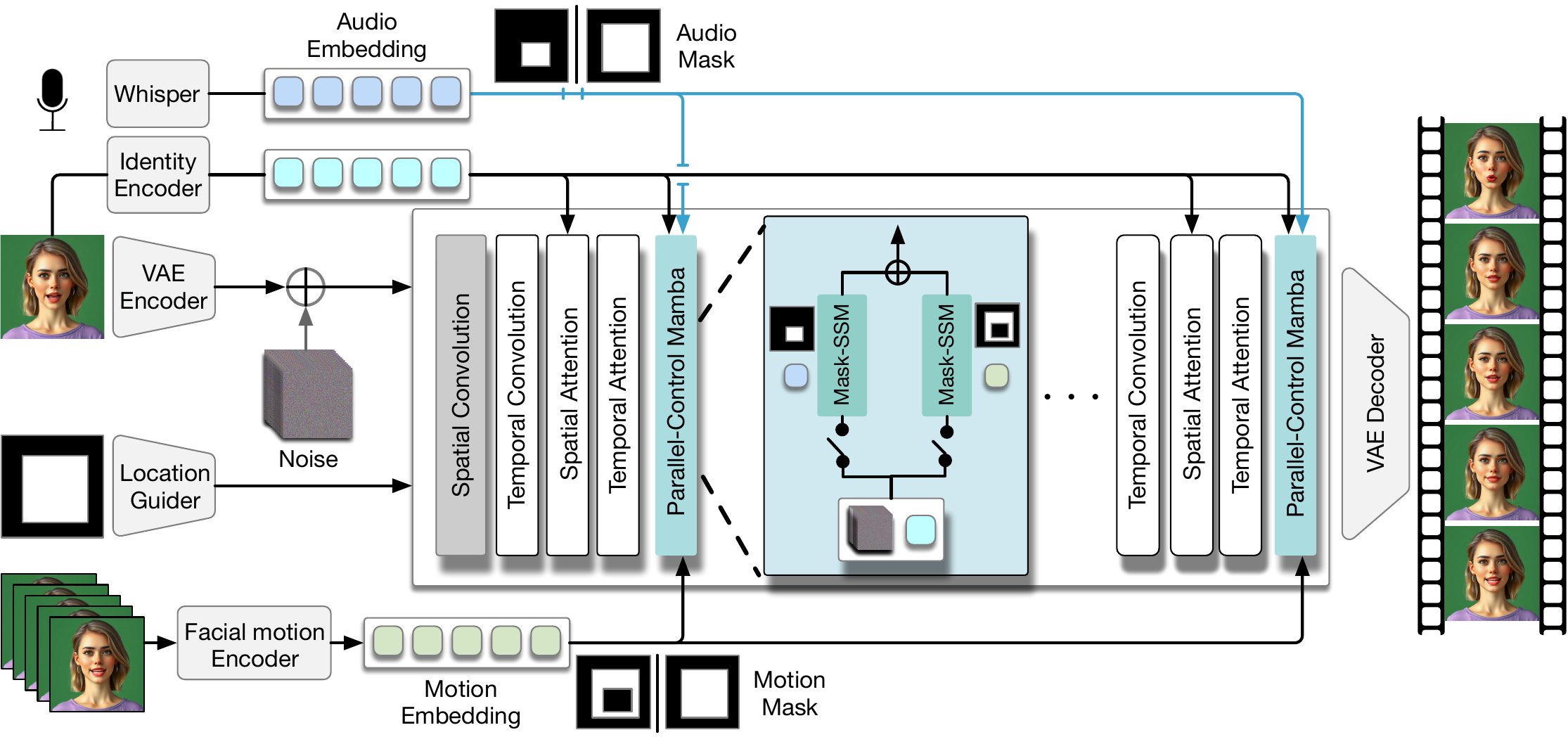}
    \vspace{-10pt}
    \caption{Illustration of our ACTalker framework. ACTalker takes multiple signals inputs (i.e., audio and visual facial motion) to drive the generation of talking head videos. In addition to the standard layers (e.g., spatial convolution, temporal convolution, spatial attention, and temporal attention) in the stable video diffusion model, we introduce a parallel-control mamba layer to harness the power of multiple signals control. Audio and facial motion signals are fed into this parallel-control mamba layer, along with their corresponding masks, which indicates the regions to focus on for manipulation.
    }
    \vspace{-10pt}
    \label{fig:framework}    
\end{figure*}
Our contributions can be summarized as follows:
\begin{itemize}
   \item We propose the audio-visual controlled video diffusion model for talking head generation, which enables seamless and simultaneous control of generated videos using both audio and fine-grained facial motion signals, leading to more realistic and expressive outputs.

    \item We introduce the parallel-control mamba layer (PCM), which effectively coordinates multiple driving signals without conflicts, ensuring smooth integration of audio and facial motion signals. Additionally, we incorporate a mask-drop strategy that directs the model’s focus to the relevant facial regions for each control signal, improving both the quality and computational efficiency of the generated video.
    
    \item We perform extensive experiments, including evaluations on challenging datasets, demonstrating that our method generates natural-looking talking head videos with precise control over multiple signals, achieving superior results in multiple signals video synthesis.
\end{itemize}

\vspace{-5pt}
\section{Related Work}
\noindent\textbf{Talking Head Generation.}
Talking head generation has been a longstanding challenge in the fields of computer vision and graphics. Recent advancements in the field of talking head generation can be divided into two subcategories: non-diffusion-based and diffusion-based methods.
Non-diffusion-based methods~\cite{ji2022eamm,gururani2023space,liu2023moda,zhang2023sadtalker} are known for their ability to achieve realistic facial animations and fidelity to motion. Some expression-driven methods~\cite{hong2023dagan++,hong2022depth,hong2023implicit} employ Taylor approximation to estimate the motion flow between two face and then warping the source image. Some audio-driven talking head methods~\cite{gururani2023space,zhang2022sadtalker} map the audio to the spatial expression landmarks and then control the facial expression and lip movement by audio following the pipeline of expression-driven methods~\cite{zhao2024synergizing}.
With the development of diffusion models, recent works~\cite{tian2025emo,wei2024aniportrait,xu2024vasa,xu2024hallo,chen2024echomimic,jin2024alignment,jiang2024loopy,ma2024follow} have adopted stable diffusion~\cite{rombach2022high} and motion modules~\cite{guo2023animatediff} for talking head generation within a two-stage training paradigm. Follow-Your-Emoji~\cite{ma2024follow} leverages landmarks as motion representations to guide video generation. X-Portrait~\cite{xie2024x} first constructs cross-identity training pairs using a pretrained talking head model, and then employs a ControlNet-style network to predict the results. Hallo~\cite{xu2024hallo} introduces a hierarchical mask that enables audio-driven control of portrait videos.

However, most previous methods only allow single-signal control at a time. Therefore, we propose a novel framework based on the mamba structure that can generate videos driven by either multiple signals or a single signal at a time. Moreover, our architectural advancements enhance the model's ability to simultaneously learn spatial and temporal relationships within the mamba structure.

\begin{figure*}[ht]
  \centering
    \includegraphics[width=1\linewidth]{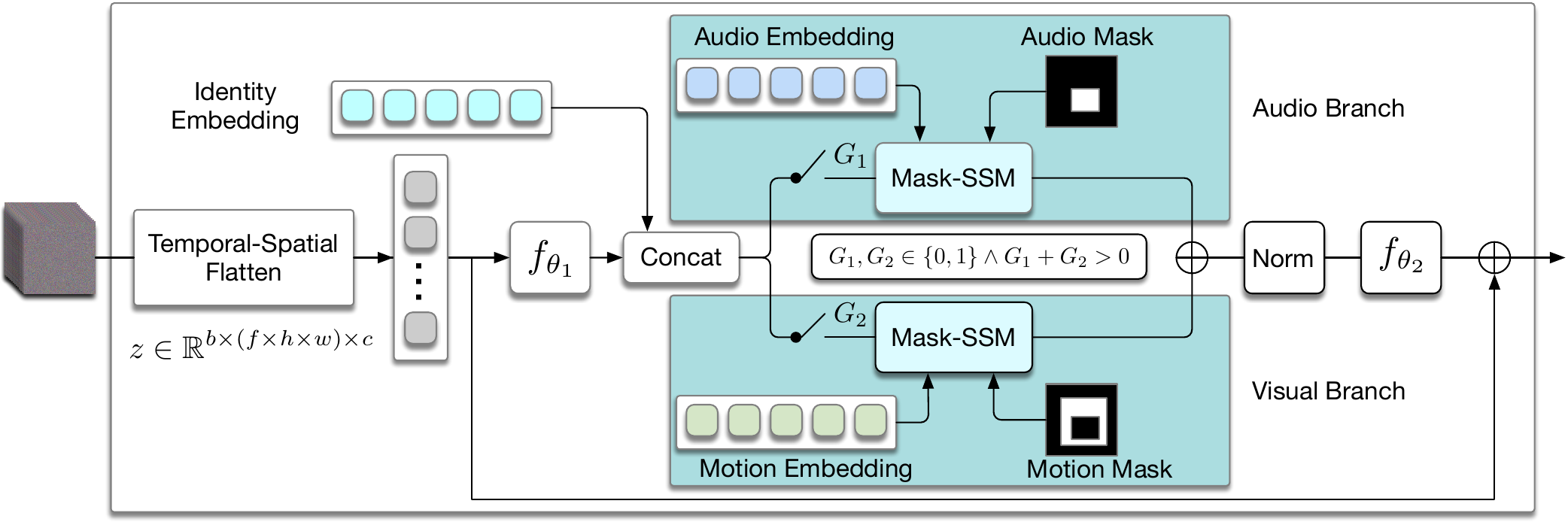}
    \vspace{-20pt}
    \caption{Illustration of parallel-control mamba layer. There are two parallel branches in this layer, one for audio control and the other is for expression control. We utilize a gate in each branch to control the accessing of control signal during training. During inference, we can manually modify the statue of gates to enable single signal control or multiple signals control.
    }
    \vspace{-10pt}
    \label{fig:mambalayer}    
\end{figure*}
\noindent\textbf{Selective State Space Models.}
State Space Models (SSMs) have recently been proposed to integrate deep learning for state space transformation~\cite{s4gu2021,fu2022hungry}. Inspired by continuous state space models in control systems, SSMs, when combined with HiPPO initialization~\cite{hippogu2020}, show great potential in addressing long-range dependency issues, as demonstrated in LSSL~\cite{lsslgu2021}. However, the computational and memory demands of state representation make LSSL impractical for real-world applications. To address this, S4~\cite{s4gu2021} introduces parameter normalization into a diagonal structure. This has led to the development of various structured SSMs with different configurations, such as complex-diagonal structures~\cite{dssgupta2022,s4dgu2022}, MIMO support~\cite{s5smith2022simplified}, diagonal-plus-low-rank decomposition~\cite{liquids4hasani2022liquid}, and selection mechanisms~\cite{mambagu2023mamba}, which have been integrated into large-scale frameworks~\cite{megama2022mega,gssmehta2023long}. Recently, SSMs have been applied to language understanding~\cite{megama2022mega,gssmehta2023long,xiao2024grootvl}, content-based reasoning~\cite{mambagu2023mamba}, motion generation~\cite{xu2024mambatalk} and one-dimensional image classification at the pixel level~\cite{s4gu2021}, leading to significant improvements.

In this work, we integrate SSMs into 2D talking head generation to address the challenge of aggregating features with control signals. By applying SSMs, our framework efficiently integrates contextual information from audio, video, and spatiotemporal features, demonstrating the potential of ACTalker for talking heads.

\section{Methodology}
In this work, we aim to develop a novel video diffusion model for one-shot talking head video generation driven by multiple signals. We propose an audio-visual controlled video diffusion model that provides flexible control over the availability of each driving signal, enabling effective multiple signals control of the generated video.
\vspace{-5pt}
\subsection{Overview} 
\vspace{-5pt}
In this work, we adopt the stable video diffusion model (SVD)~\cite{blattmann2023stable} as our codebase. As shown in Figure~\ref{fig:framework}, in addition to the regular inputs of the source image and pose image, our ACTalker also accepts multiple driving signals, such as audio and visual expressions, to guide video generation. Given a source image $\textbf{I}_s$, a face mask image $\mathcal{M}_{face}$, facial motion sequences $\{\textbf{I}_{exp}^i\}_{i=1}^N$, and an audio segment $\mathcal{A}$, we first use a VAE encoder to encode the source image into latent space, which is then concatenated with noise latent. Next, we use Whisper~\cite{radford2023robust} to extract the audio embedding $\mathbf{e}_a$ from the audio $\mathcal{A}$. Similarly, we utilize a pre-trained motion encoder~\cite{xu2024vasa} to extract an implicit facial motion embedding $\mathbf{e}_{mtn}$ from a sequence of face images, and an identity encoder~\cite{deng2019arcface} to obtain the identity embedding $\mathbf{e}_{id}$ from the source image $\textbf{I}_s$. 

In addition to the regular layers in SVD, such as spatial convolution, temporal convolution, appearance attention, and temporal attention, we design a parallel-control mamba layer (PCM) in each block to enable multi-signal control. The PCM layer consists of multiple branches, each containing a Mask-SSM unit. Specifically, in each branch, the Mask-SSM takes one driven signal and its corresponding mask as input to manipulate the selected spatial-temporal feature tokens by aggregating them in the SSM structure, achieving facial control. Additionally, a gate mechanism is used in the PCM to manage the control of each driven signal. Each branch maintains a gate to decide the availability of the corresponding branch. 
\vspace{-5pt}
\subsection{Parallel-control Mamba Layer}
\vspace{-5pt}
In this work, we aim to develop a method for generating portrait videos controlled by either multiple signals without conflict or a single signal. To this end, we propose a novel parallel-control Mamba layer (Figure~\ref{fig:mambalayer}) that leverages driven signals to manipulate the temporal-spatial features via the Mamba structure, achieving fine-grained control over facial synthesis. The layer consists of two primary branches, each controlling different facial regions with different conditioning signals: audio and facial motion. Additionally, we designed a gate mechanism to achieve flexible control by controlling the activation of each branch.

\noindent\textbf{Identity Preservation.} As illustrated in Figure~\ref{fig:mambalayer}, to enable the driving signal to influence the intermediate noise feature both spatially and temporally, we first flatten the spatiotemporal intermediate feature across its spatial and temporal dimensions. This produces a flattened feature $z \in \mathbb{R}^{b \times (f \times h \times w) \times c}$, where $f$ represents the number of frames, and $h$ and $ w$ are the height and width of the original spatiotemporal feature. Furthermore, to preserve identity during face manipulation driven by the signal, we also aggregate the identity embedding $ \mathbf{e}_{id}$ with the noise feature:
\begin{equation}
    z' = \text{Concat}(\mathbf{e}_{id}, f_{\theta_1}(z)), \quad z' \in \mathbb{R}^{b \times n_1 \times c},
    \label{eq:concate-id}
\end{equation}
where $f_{\theta_1}$, parameterized by $\theta_1$, is an MLP that transforms the noise feature to integrate with the identity embedding.

\noindent\textbf{Multiple Signals Control with Gate Mechanism.}  
To enable multiple signals control, we feed the concatenated feature $z'$ into parallel branches, where each branch is responsible for controlling a specific facial region. As shown in Figure~\ref{fig:mambalayer}, we have two branches to handle audio-driven and motion-driven control, respectively. In our setup, we expect our model to generate portrait videos driven by both signals simultaneously, while still maintaining the capability to be driven by a single signals. To achieve this, we randomly set up gate variables in each branch ($G_1$ and $ G_2$) to control the driving mode during training. For the gates in both branches, we impose the constraint:
\begin{equation}
    G_1, G_2 \in \{0,1\} \quad \land \quad G_1 + G_2 > 0,
\end{equation}

There are three possible configurations under this constraint: $\{G_1 = 0, G_2 = 1\}$, $\{ G_1 = 1, G_2 = 0 \}$, and $\{G_1 = 1, G_2 = 1 \}$. The first two configurations mean that only one signal is used for control during training, while the last configuration indicates that both signals are used simultaneously. During our training, we randomly select one of these three gate statuses. Therefore, our model is able to generate portrait video under the single signals or both audio-visual signals control.

\noindent\textbf{Multi-control Aggregation.} For the outputs of each branch ($\overline{z}_1$ and $\overline{z}_2$, with the branch structure detailed later), we concatenate them and apply normalization to the concatenated results in order to improve training stability:
\begin{equation}
    o_1 = \text{Norm} (\overline{z}_1 + \overline{z}_2).
\end{equation}

Next, we apply a residual connection between the aggregated output $ o_{\text{agg}}$ and the original flattened noise feature $ z$:
\begin{equation}
    o_2 = f_{\theta_2}(o_1) + z,
\end{equation}
where $ f_{\theta_2}$ is an MLP that transforms the aggregated feature $ o_1$, parameterized by $ \theta_2$. Finally, the entire parallel-control mamba layer outputs $ o_2$, which is passed to the next block in our framework.

The audio and motion-manipulated features are then aggregated and pass the control signal information throughout the framework without conflicts.
\subsection{Mask-SSM}
\begin{figure}[t]
  \centering
    \includegraphics[width=1\linewidth]{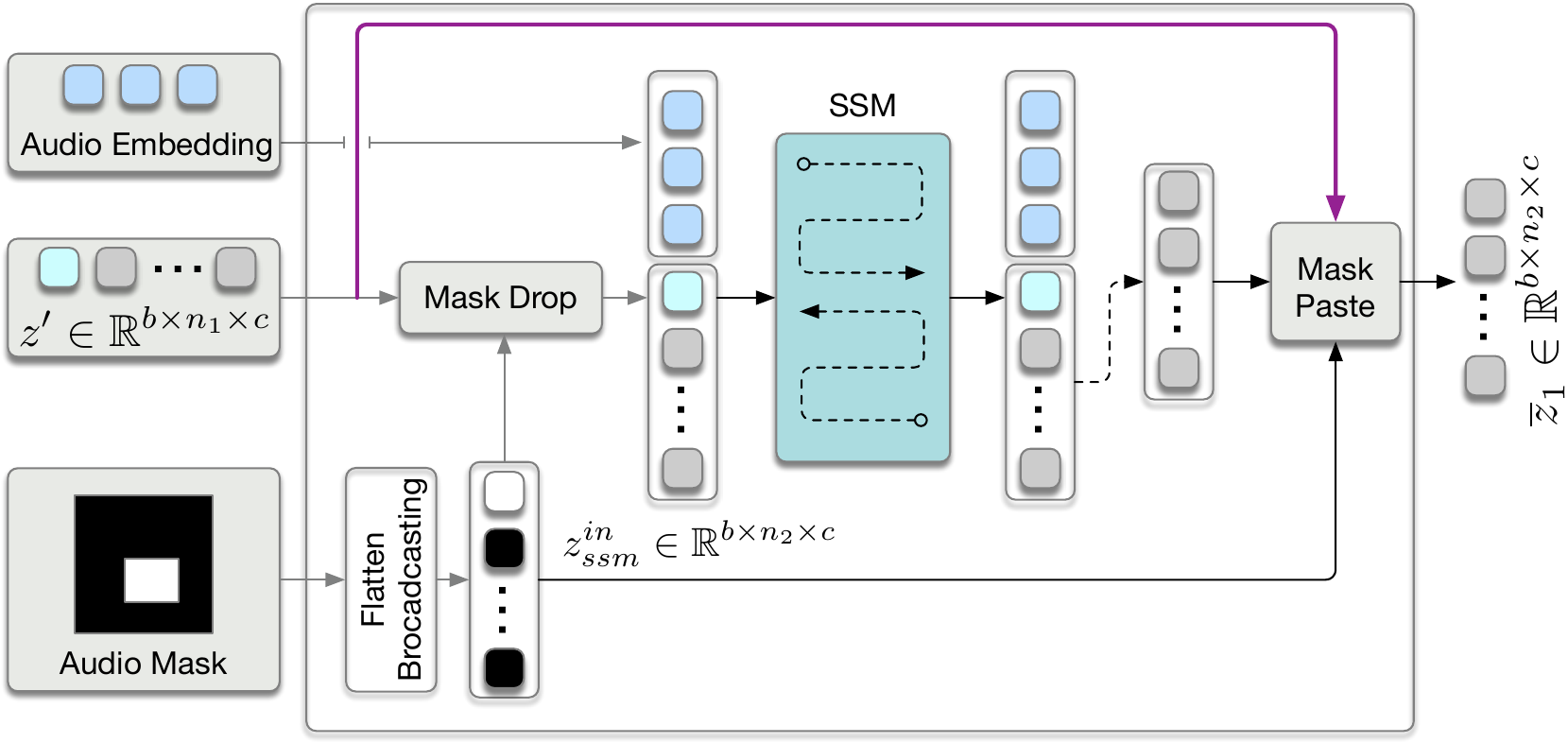}
    \vspace{-20pt}
    \caption{The illustrating of the Mask-SSM in audio branch of parallel-control mamba layer. The visual branch is the same but replace with the motion embedding and motion mask
    }
    \vspace{-10pt}
    \label{fig:ssm}    
\end{figure}
As shown in Figure~\ref{fig:mambalayer}, in each branch, we design a Mask State Space Model (Mask-SSM) to process the input spatiotemporal noise feature and control signal. 
Since we flatten the noise feature volume along both the spatial and temporal dimensions, the number of tokens increases dramatically (\(\textbf{frames} \times \textbf{width} \times \textbf{height}\)) becomes much larger in the shallow layers). 
To efficiently fuse each token with the driving signal, we adopt the state space model in our framework. To solve the control conflict problem, we design a specific mask (see \emph{Supplementary Material}) for each driven signal to indicate their control regions.
Based on that mask, we design a mask-drop strategy to not only reduce the number of noise feature tokens by dropping irrelevant ones but also distinguish the tokens across spatial and temporal dimensions that need to be manipulated by the corresponding signals, as indicated by the input mask. The mask-drop strategy mainly consists of two steps: Mask Drop and Mask Paste. Each branch in PCM shares the same architecture and specifies the control region of the driven signals using different masks, \ie, the audio mask and the motion mask. In this section, we take the audio branch as an example to demonstrate the details.
\begin{figure*}[t]
  \centering
    \includegraphics[width=1\linewidth]{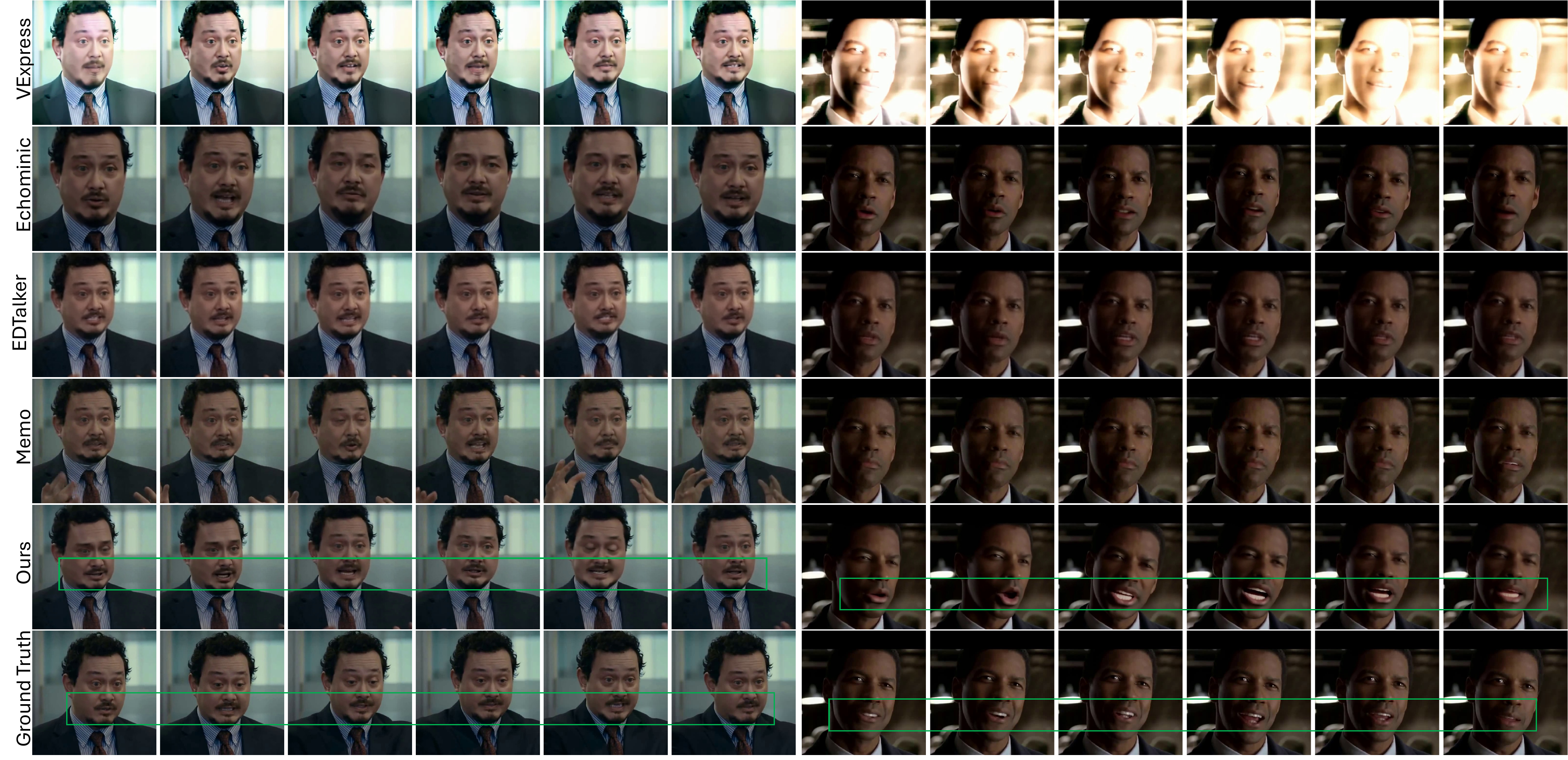}
    \vspace{-20pt}
    \caption{Comparison of different methods for audio-driven talking head generation. Our method can produce more natural and accurate lip-synced videos. Due to the page limitation, the results of SadTalker~\cite{zhang2022sadtalker} and Hallo~\cite{xu2024hallo} are reported in \emph{Supplementary Material} 
    }
    \vspace{-10pt}
    \label{fig:audio-comparision}  
\end{figure*}
\noindent\textbf{Mask Drop.} As shown in Figure~\ref{fig:ssm}, given an audio mask $ \mathcal{M}_{audio}$, where the control region is set to 1 and all other regions are set to 0, we first flatten the mask and broadcast it to match the shape of the input noise feature $z'$. Then, we apply this flattened mask to drop the noise features (we omit the identity tokens concatenated in Eq.~\ref{eq:concate-id} for simplicity). This process is expressed as:
\begin{equation}
    z_{ssm}^{in} = \mathcal{D}(z', \mathcal{M}_{audio}),
\end{equation}
where $ z_{ssm}^{in} \in \mathbb{R}^{b \times n_2 \times c}$, $ z' \in \mathbb{R}^{b \times n_1 \times c}$, and $ n_1 > n_2$. Here,  $\mathcal{D}$ represents the drop operation, which removes the tokens where the corresponding position in the audio mask $ \mathcal{M}_{audio}$ is zero.

\noindent\textbf{Mask Paste.} After obtaining the masked tokens $z_{ssm}^{in}$, we concatenate them with the driving signal, i.e., the audio embedding, and then pass the concatenated result into an SSM unit to enable each token to interact with the driving signal:
\begin{equation}
    z_{ssm}^{out} = SSM(Concat(z_{ssm}^{in}, \mathbf{e}_a)).
\end{equation}
Consequently, we drop the audio and identity tokens in the result $ z_{ssm}^{out}$. The resulting tokens only maintain the facial information of the controlled region after processing by the driven signal. To cooperate with other regions, such as the background, we paste the audio-aggregated tokens back to the original noise feature $z$ according to the mask $ \mathcal{M}$:
\begin{equation}
    \overline{z}_1 \leftarrow z[\mathcal{M}_{audio} == 1] = z_{ssm}^{out}.
    \label{eq:mask-paste}
\end{equation}
Therefore, we replace the tokens of the control region in $z$ with the aggregated feature tokens $z_{ssm}^{out}$ to obtain the audio-aggregated feature $\overline{z}_1$. We can also get the motion-aggregated feature $\overline{z}_2$ in the same way but in a different Mask-SSM branch.

In this way, Mask-SSM can resolve control conflicts by distributing different tokens to each driven signal using the mask-drop strategy. By applying the mask-drop strategy in each control branch, we can aggregate features spatio-temporally with reduced computational complexity through the Mamba structure while improving the model's focus on signal-specific regions of the face, leading to more accurate control of video generation.

\subsection{Training and Inference}
\noindent\textbf{Training.} To train our video diffusion model, we apply the general training objective of the video diffusion model:
\begin{equation}
    \mathcal{L} = \mathbb{E}_{t,z,\epsilon}[||\epsilon - \epsilon_{\theta}(\mathcal{C},z, t)||],
\end{equation}
where $z$ denotes the latent embedding of the training sample, $ \epsilon$ and $ \epsilon_\theta$ are the ground truth noise at the corresponding timestep $ t$ and the predicted noise by our ACTalker, respectively. $\mathcal{C}$ is the condition set, which includes the audio embedding, motion embedding, identity embedding, and pose location. During training, we randomly select one of the three gate configurations in our parallel-control mamba layer to enable flexible controllability. In the step of training the model with single-signal control, we repurpose the pose mask $\mathbf{I}_p$ as an audio/expression mask, allowing the driving signal to control the entire face.

\noindent\textbf{Inference.} During inference, we manually set the gate status to enable different types of control (audio-only, expression-only, and audio-expression). Our ACTalker is capable of generating videos of arbitrary length within memory constraints, given reference images and audio or facial motion driving signals. We also apply classifier-free guidance (CFG)~\cite{ho2022classifier} to achieve better results.

\begin{table*}[t]
  \centering
  \resizebox{1\linewidth}{!}{
        \begin{tabular}{lcccccc|cccccc}
        \toprule
          \textbf{Model} & \textbf{Sync-C$\uparrow$} & \textbf{Sync-D $\downarrow$} & \textbf{FVD-Res $\downarrow$} & \textbf{FVD-Inc $\downarrow$} & \textbf{FID $\downarrow$} & \textbf{Smooth $\uparrow$} & \textbf{Sync-C$\uparrow$} & \textbf{Sync-D $\downarrow$} & \textbf{FVD-Res $\downarrow$} & \textbf{FVD-Inc $\downarrow$} & \textbf{FID $\downarrow$} & \textbf{Smooth $\uparrow$} \\
          \toprule
        SadTalker\cite{zhang2022sadtalker} & 3.814 & 8.824 & 18.484 & 352.296 & 51.804 & 0.9963 & 3.899 & 7.895 & 16.642 & 264.065 & 44.965 & 0.9953 \\
        Hallo\cite{xu2024hallo} & 4.316 & 9.020 &13.317 & 342.965& 37.400 & 0.9946  & 3.963 & 8.125 & 6.888 & 266.920 & 23.157 & 0.9941 \\
        VExpress\cite{wang2024v} & 3.612 & 9.165 & 37.657 & 539.920 & 58.427 & 0.9959 & 4.888 & 7.898 & 14.950 & 517.880 & 26.753 & 0.9954 \\
        EDTalk~\cite{tan2024edtalk} & 5.124 & 8.438 & 16.723 & 430.906 & 50.428 & 0.9972 & 4.759 & 8.375 & 14.114 & 477.147& 50.135 & 0.9954 \\
        EchoMimic~\cite{chen2024echomimic} & 2.989 & 10.188 & 16.897 & 366.007 & 45.489 & 0.9938 & 3.239 & 9.411 & 46.038 & 450.798 & 41.357 & 0.9923 \\
            Memo~\cite{zheng2024memo} & 3.958 & 9.118& 7.992& 264.596 & 31.134 & 0.9954 & 5.093 & 7.854 & 5.098 & 194.570 & 18.837 & 0.9945 \\
            Ours (Only Audio) & \textbf{5.317} & \textbf{7.869} & \textbf{7.328} & \textbf{232.374} & \textbf{30.721} & \textbf{0.9978} & \textbf{5.334} & \textbf{7.569} & \textbf{4.754} & \textbf{193.120} & \textbf{16.730} & \textbf{0.9955} \\
            \rowcolor{lightgray}Ours (Audio-Visual) & \textbf{5.737} & \textbf{7.510 }& \textbf{7.074} & \textbf{230.125} & \textbf{29.977} & \textbf{0.9979} & \textbf{5.511} & \textbf{7.311} & \textbf{4.574} & \textbf{190.125} & \textbf{15.977} & \textbf{0.9955} \\
        \bottomrule
        \end{tabular}
        }
        \vspace{-10pt}
\caption{Audio-driven comparison of different methods on Celebv-HQ dataset (left) and RAVDESS dataset (right).}
        \vspace{-10pt}
\label{tab:audio}
\end{table*}
        
\section{Experiments}
In this section, we present both quantitative and qualitative experiments to validate the effectiveness of ACTalker. More implementation details, additional results and video demonstrations can be found in the \emph{Supplementary Materials}. We strongly recommend watching the video demos.

\begin{figure*}[t]
  \centering
    \includegraphics[width=0.97\linewidth]{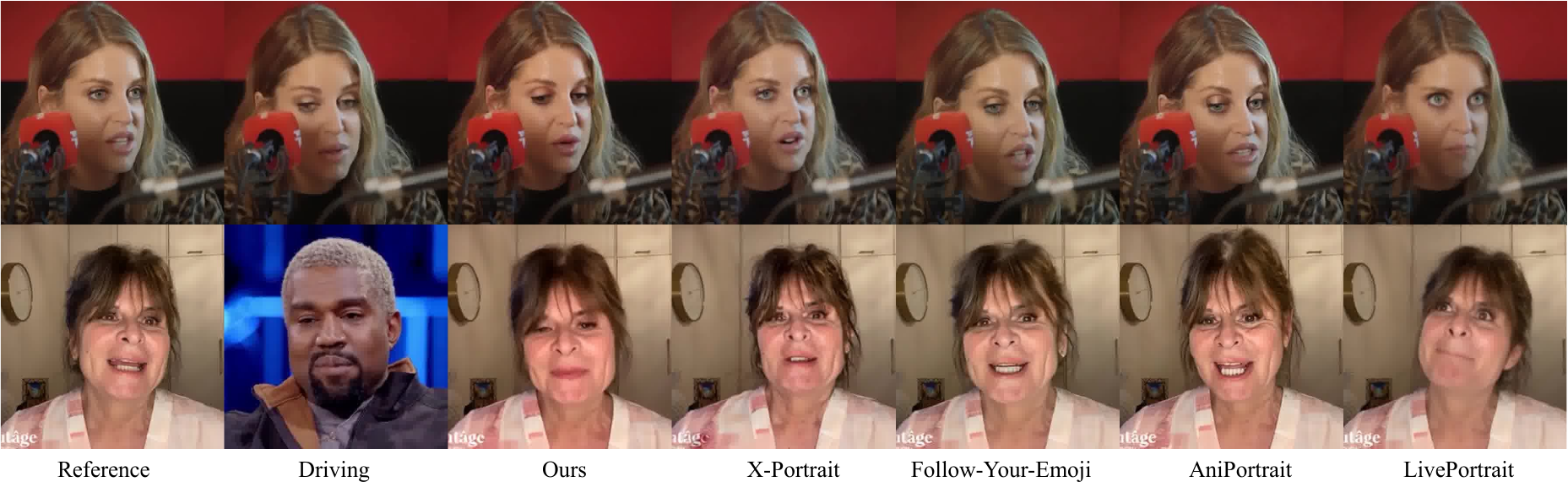}
    \vspace{-15pt}
    \caption{Comparison of different methods on VFHQ. Self reenactment (first row) and cross reenactment (last row).
    }
    \vspace{-8pt}
    \label{fig:face-comparision}    
\end{figure*}

\begin{table*}[t]
\centering
\resizebox{1\linewidth}{!}{
\begin{tabular}{lcccccc|cccc}
\toprule
\textbf{Model} & \textbf{LMD($\times 10^{-2}$) $\downarrow$} & \textbf{FID $\downarrow$} & \textbf{FVD-Inc $\downarrow$} & \textbf{PSNR $\uparrow$} & \textbf{SSIM $\uparrow$} & \textbf{LPIPS $\downarrow$} & \textbf{Pose Distance $\downarrow$} & \textbf{Expression Similarity $\uparrow$} & \textbf{ID Similarity($\times 10^{-1}$)$\uparrow$} & \textbf{Smooth ($\times 10^{-2}$)$\uparrow$} \\
\midrule
LivePortrait~\cite{guo2024liveportrait} & {\large 0.49} & {\large 82.69} & {\large 483.42} & {\large 40.37} & {\large 0.92} & {\large 0.31} & {\large 26.99} & {\large 0.38} & {\large 8.55} & \textbf{{\large 99.53}} \\
AniPortrait~\cite{wei2024aniportrait}   & {\large 0.68} & {\large 81.89} & {\large 430.27} & {\large 39.29} & {\large 0.85} & {\large 0.36} & {\large 21.31} & {\large 0.46} & {\large 8.50} & {\large 99.36} \\
FollowYourEmoji~\cite{ma2024follow}      & {\large 0.65} & {\large 77.17} & {\large 417.51} & {\large 39.67} & {\large 0.86} & {\large 0.35} & {\large 20.94} & {\large 0.48} & {\large 8.59} & {\large 98.99} \\
X-Portrait~\cite{xie2024x}                & {\large 0.24} & {\large 82.92} & {\large 416.42} & {\large 39.64} & {\large 0.92} & {\large 0.27} & {\large 20.38} & {\large 0.48} & {\large 8.57} & {\large 99.39} \\
\rowcolor{lightgray}Ours                                     & \textbf{{\large 0.14}} & \textbf{{\large 75.47}} & \textbf{{\large 358.82}} & \textbf{{\large 40.65}} & \textbf{{\large 0.94}} & \textbf{{\large 0.24}} & \textbf{{\large 20.32}} & \textbf{{\large 0.57}} & \textbf{{\large 8.64}} & {\large 99.48} \\
\bottomrule
\end{tabular}
}
\vspace{-10pt}
\caption{
Comparison of different methods on VFHQ. Self reenactment (left) and cross reenactment (right). 
}
\vspace{-10pt}
\label{tab:face}
\end{table*}
\subsection{Experimental Settings}
\noindent\textbf{Dataset.} We use publicly available datasets, such as HDTF~\cite{zhang2021HDTF}, VFHQ~\cite{xie2022vfhq}, VoxCeleb2~\cite{chung2018voxceleb2}, CelebV-Text~\cite{yu2022celebvtext}, along with self-collected videos, to create a diverse training dataset. Since our method is capable of both audio-driven talking head generation and expression-driven face reenactment, we conduct comparisons on these two tasks. For audio-driven talking head generation, we follow the settings of Loopy~\cite{jiang2024loopy}, sampling 100 videos from CelebV-HQ~\cite{zhu2022celebv} and RAVDESS (Kaggle). Additionally, we test our face reenactment capabilities using the VFHQ dataset~\cite{xie2022vfhq}. 

\subsection{Quantitative and Qualitative Analysis}
We conduct a quantitative and qualitative comparison with other methods on audio-driven talking head generation and face reenactment tasks. During inference, we set the gate value to either 0 or 1 to control whether the video is generated using the corresponding signal. The quantitative results are reported in Table.~\ref{tab:audio} and Table.~\ref{tab:face}. Also, we visualize the qualitative comparison in Figure.~\ref{fig:audio-comparision} and Figure.~\ref{fig:face-comparision}. 

\noindent\textbf{Audio-driven Talking Head Results.} We first compare our method with other audio-driven talking head generation methods. The results reported in Table~\ref{tab:audio} and Table~\ref{tab:audio} strongly demonstrate the superiority of our approach compared with existing works. Our method achieves the best Sync-C and Sync-D scores on the CelebV-HD dataset ($5.317$ for Sync-C and $7.869$ for Sync-D), verifying that it can produce audio-synchronized talking head videos. In terms of video quality, our method also shows significant improvement. For example, our approach obtains an FVD-Inc score of $232.374$, outperforming the second-best method Memo~\cite{zheng2024memo} by roughly $32$ points. These results demonstrate that our specific design in the stable video diffusion model brings notable benefits. Additionally, Figure~\ref{fig:audio-comparision} visualizes several samples of audio-driven talking head generation. It can be observed that our results exhibit accurate lip motion and fewer artifacts compared with other methods. These findings confirm that our mamba structure design is beneficial for lip-sync by directly manipulating the selected tokens.

\noindent\textbf{Face Reenactment.} In addition to audio-driven talking head video generation, our framework is also capable of performing expression-driven talking head video generation, i.e., face reenactment. Existing methods~\cite{hong2022depth,hong2023implicit} typically evaluate face reenactment in two scenarios: self-reenactment and cross-reenactment. In self-reenactment, the driving video and reference share the same identity, whereas in cross-reenactment they have different identities. As shown in Table~\ref{tab:face}, our expression-driven method achieves superior results compared with existing state-of-the-art approaches. Specifically, our method outperforms X-Portrait~\cite{xie2024x} by $9\%$ in expression similarity during cross reenactment, while also maintaining the best ID similarity (8.64 for cross reenactment). These results validate that our mamba structure effectively manipulates facial region tokens to perform accurate facial expression animation. We further visualize sample outputs in Figure~\ref{fig:face-comparision}. Compared with other methods, our approach captures more subtle micro-motions—such as the mouth movements in the first two self-reenactment samples—and produces more precise expression animations in the last two cross-reenactment samples. It verifies that our designed Mask-SSM effectively enhances the generated content in the controlled region.

\begin{figure}[t]
  \centering
    \includegraphics[width=1\linewidth]{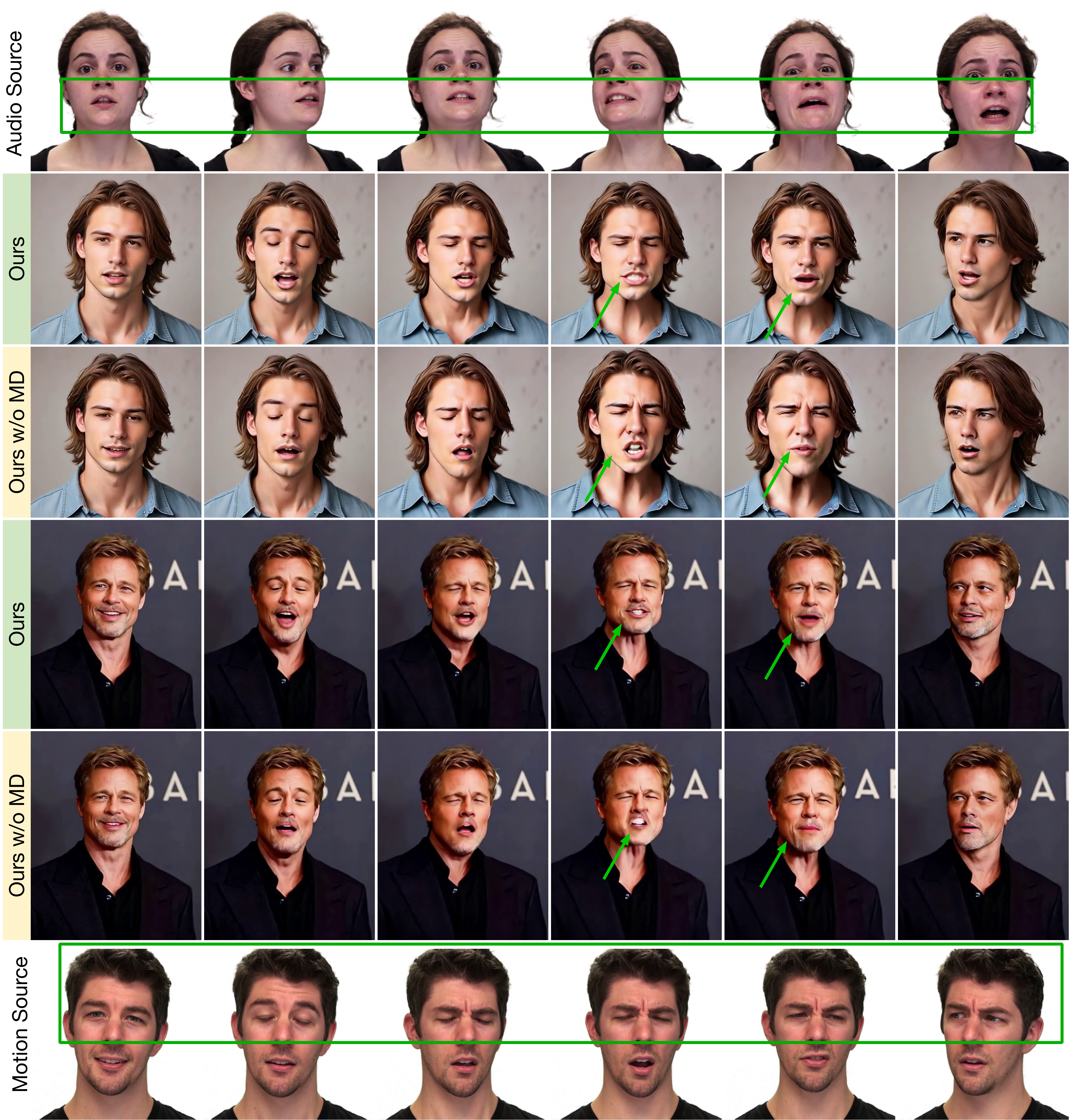}
    \vspace{-15pt}
    \caption{Visualization of multiple signals control. Our generated video accurately replicates the lip movements driven by the audio source and captures the head motion—particularly the eye movements and pose—as guided by the motion source. Once we remove the masks in both Mask-SSMs and generate the video using multiple driving signals, the motion source can also affect the mouth movement (``Ours w/o MD''), causing a control conflict.
    }
    \vspace{-10pt}
    \label{fig:multimodal}    
\end{figure}

\begin{figure}[t]
  \centering
    \includegraphics[width=1\linewidth]{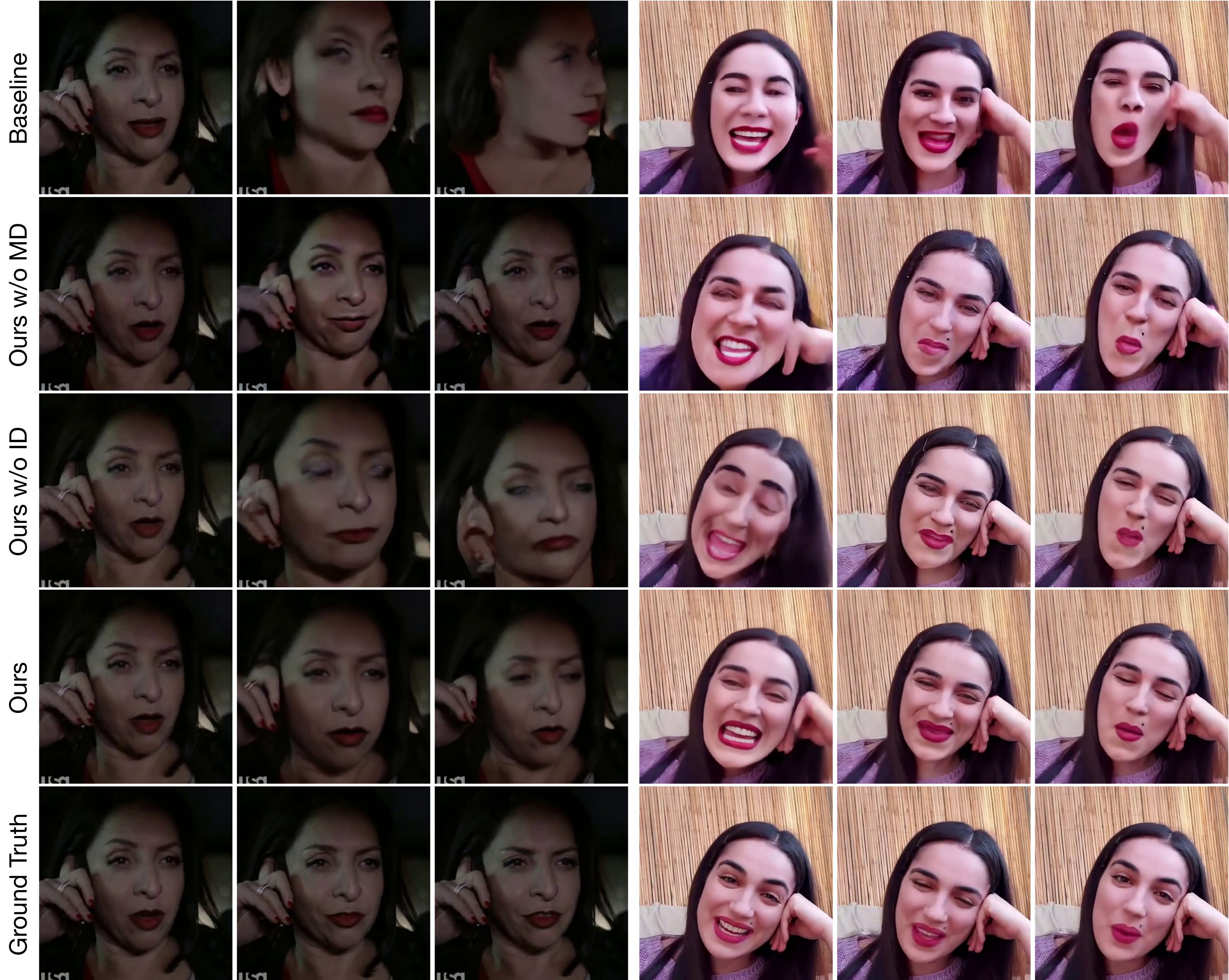}
    \vspace{-10pt}
    \caption{The visualization of ablation studies driven by audio. Our full method can produce more natural videos.
    }
    \vspace{-10pt}
    \label{fig:abaltion}    
\end{figure}
\begin{table}[t]
  \centering
  \resizebox{1\linewidth}{!}{
        \begin{tabular}{lcccccc}
        \toprule
          \textbf{Model} & \textbf{Sync-C$\uparrow$} & \textbf{Sync-D $\downarrow$} & \textbf{FVD-Res $\downarrow$} & \textbf{FVD-Inc $\downarrow$} & \textbf{FID $\downarrow$} & \textbf{Smooth $\uparrow$} \\
        \toprule
         Baseline & {\large 4.592} & {\large 8.523} & {\large 16.983} & {\large 268.512} & {\large 32.483} & {\large 0.9967} \\
        Ours w/o MD & {\large 4.953} & {\large 8.184} & {\large 7.456} & {\large 240.651} & {\large 31.268} & {\large 0.9969} \\ 
        Ours w/o ID & {\large 5.241} & {\large 7.748} & {\large 8.364} & {\large 247.933} & {\large 31.170} & {\large 0.9978} \\
         Ours (Only Audio) & \textbf{{\large 5.317}} & \textbf{{\large 7.869}} & \textbf{{\large 7.328}} & \textbf{{\large 232.374}} & \textbf{{\large 30.721}} & \textbf{{\large 0.9978}} \\
         \rowcolor{lightgray}Ours (Audio-Visual) & \textbf{{\large 5.737}} & \textbf{{\large 7.510}} & \textbf{{\large 7.074}} & \textbf{{\large 230.125}} & \textbf{{\large 29.977}} & \textbf{{\large 0.9979}} \\
        \bottomrule
        \end{tabular}
  }
  \vspace{-10pt}
\caption{Ablation studies on Celebv-HQ dataset for audio-driven.}
\label{tab:ab-audio-C}
\end{table}
\subsection{Ablation Study}
In this section, we evaluate each design in our framework to verify its effectiveness. The results are reported in Table~\ref{tab:ab-audio-C}, Figure~\ref{fig:multimodal},  and Figure~\ref{fig:abaltion}.

\noindent\textbf{Multiple Signals Control.} Benefiting from our gating mechanism and Mask-SSM, our framework can generate videos driven either by a single signal (audio or expression) or by multiple signals simultaneously. We conduct multiple signals driven experiments as part of our ablation studies. As shown in Figure~\ref{fig:multimodal}, we generate three samples using the same audio and expression signals. We observe that the generated videos exhibit consistent expressions (e.g., similar eye status) and synchronized mouth movements. Quantitative results further demonstrate that videos driven by both signals yield superior performance compared to those generated using a single signal. These findings confirm that our parallel-control mamba layer effectively enables different signals to control disentangled facial regions, achieving robust multiple signals control.

\noindent\textbf{Mamba Structure. }To evaluate the effectiveness of our mamba structure, we integrate it into the Stable Video Diffusion model and construct a baseline by replacing the parallel-control mamba layer with a spatial cross-attention layer. As shown in Table~\ref{tab:ab-audio-C} and Figure~\ref{fig:abaltion}, without our mamba structure, both lip synchronization and overall video quality deteriorate dramatically. These results confirm that our mamba structure effectively captures the core information from the driving signal and broadcasts it across temporal and spatial dimensions, resulting in more natural portrait video generation.

\noindent\textbf{Mask-Drop and Control Conflict.} We employ a mask-drop strategy in our mamba structure not only to reduce the number of processed tokens but also to enhance the focus of the driving signal on the control regions. We conduct an ablation study (labeled ``Ours w/o MD'' in Table~\ref{tab:ab-audio-C} and illustrated in Figure~\ref{fig:abaltion}) to verify its effectiveness. As shown in Table~\ref{tab:ab-audio-C}, without the mask-drop strategy, the model is distracted by irrelevant tokens, resulting in a performance drop (the Sync-C score is $4.953$ compared to $5.317$ with the full method). Moreover, the generated outputs appear less natural without the mask-drop strategy. Also, we show the samples that are driven by both signals in Figure~\ref{fig:multimodal}. We can observe that, without the mask-drop strategy, the mouth region is affected by the motion signals, which is not what we expected. These results confirm that the mask-drop strategy significantly improves the controllability of the driving signal over the target regions and resolves the control conflict.

\noindent\textbf{Identity Embedding in PCM}. In our parallel-control mamba layer (PCM), we inject an identity embedding and aggregate it within the Mask-SSM to preserve identity while manipulating the selected tokens. We also perform an ablation study by removing the identity embedding from the PCM (reported as “Ours w/o ID” in Table~\ref{tab:ab-audio-C} and Figure~\ref{fig:abaltion}). As shown in Figure~\ref{fig:abaltion}, without the identity embedding, some frames fail to maintain the subject's identity, resulting in poorer quantitative performance. These findings underscore the necessity of identity embedding in our PCM layer.

\section{Conclusion}
In this work, we introduce the audio-visual controlled video diffusion (ACTalker) model, a novel end-to-end framework for talking head generation that achieves seamless and simultaneous control using both audio and fine-grained expression signals. 
Our method leverages a parallel-control mamba (PCM) layer to effectively integrate multiple driving modalities without conflict. By incorporating a mask-drop strategy, the model can focus on the relevant facial regions for each control signal, thereby enhancing video quality and preventing control conflicts in the generated videos. Extensive experiments on challenging datasets demonstrate that our approach produces natural-looking talking head videos with precise multiple signals control, achieving superior results compared to existing methods. Ablation studies verify the effectiveness of our mask-drop strategy in enhancing generated content and the gating mechanism in providing flexible control over the video generation process.
\newpage
{
    \small
    \bibliographystyle{ieeenat_fullname}
    \bibliography{main}
}
\cleardoublepage
\appendix
\section{Experiment Detail}
\subsection{Implementation.} During the training process, we resize all images and videos to $640 \times 640$. To optimize the framework, we use the AdamW optimizer with a learning rate of $1 \times 10^{-5}$. The Identity encoder~\cite{deng2019arcface} and VAE~\cite{kingma2014vae} are kept fixed, with their weights initialized from Stable Video Diffusion~\cite{blattmann2023stable}. During training, we randomly select the gate states in the parallel-control mamba layer and manually set them during inference to enable flexible control.
\subsection{Metrics}
In this work, we evaluate our method and compare it with other approaches using comprehensive quantitative metrics. Our evaluation framework consists of three main categories: (1) audio-visual synchronization, (2) visual quality assessment, and (3) facial motion accuracy. Additionally, we assess temporal smoothness and identity preservation through specialized measures.


\noindent\textbf{Audio-Visual Synchronization.} 
We employ \textbf{Sync-C} (synchronization confidence) and \textbf{Sync-D} (synchronization distance) metrics from wav2lip~\cite{prajwal2020lip} using a pretrained SyncNet~\cite{chung2017out}. Sync-C measures the confidence level of lip-audio alignment through classifier outputs, where higher values indicate better synchronization. Sync-D calculates the L2 distance between audio and visual features, with lower values representing superior alignment.

\noindent\textbf{Visual Quality Assessment.} 
We utilize four complementary metrics:
\begin{itemize}

\item \textbf{PSNR}: Peak Signal-to-Noise Ratio quantifies pixel-level fidelity through a logarithmic decibel scale, where higher values reflect better reconstruction accuracy.

\item \textbf{SSIM}~\cite{wang2004image}: Structural Similarity Index measures structural information preservation between generated and reference frames, ranging from 0 to 1, with higher values indicating better quality.
       
\item \textbf{LPIPS}~\cite{zhang2018unreasonable}: Learned Perceptual Image Patch Similarity evaluates perceptual differences using VGG~\cite{simonyan2014very} features:
\begin{equation}
\text{LPIPS}(I_{gt}^{t}, I_{gen}^{t}) = \sum_{l} w_l \left\| \mathbf{F}_l(I_{gt}^{t}) - \mathbf{F}_l(I_{gen}^{t}) \right\|_2
\end{equation}

\item \textbf{Fr\'{e}chet Inception Distance (FID)}~\cite{heusel2017gans}:  Measures feature distribution similarity between generated and real images using Inception-v3 features, with lower scores indicating better perceptual quality.
 
\item \textbf{Fr\'{e}chet Video Distance (FVD)}~\cite{unterthiner2018towards}: Assesses temporal coherence through pretrained network features:
\begin{equation}
\begin{aligned}
\text{FVD} &= \left\| \mu_{gen} - \mu_{gt} \right\|^2 + \\
&\text{Tr} \left( \Sigma_{gen} + \Sigma_{gt} - 2 \left( \Sigma_{gen} \Sigma_{gt} \right)^{1/2} \right)
\end{aligned}
\end{equation}
    
\end{itemize}

\noindent\textbf{Facial Motion Accuracy.}
For expression and pose evaluation:
\begin{itemize}
\item \textbf{Landmark Mean Distance (LMD)}: Computes the average L2 distance between facial landmarks~\cite{lugaresi2019mediapipe} of generated and reference frames, with lower values indicating better geometric accuracy.
    
\item \textbf{Pose Distance}: Measures head pose discrepancies using EMOCA~\cite{danvevcek2022emoca}-derived parameters through the mean L1 distance between generated and driving frames.
    
\item \textbf{Expression Similarity}: Calculates the cosine similarity of expression parameters from EMOCA~\cite{danvevcek2022emoca}, with higher values indicating better emotional consistency.
\end{itemize}

\noindent\textbf{Identity Similarity.} 
We employ ArcFace~\cite{deng2019arcface} scores to measure identity similarity between generated frames and reference images through deep face recognition features, where higher scores indicate better identity preservation.

\noindent\textbf{Temporal Smoothness.} 
We evaluate motion temporal smoothness by computing the optical flow consistency using VBench metrics~\cite{huang2024vbench}, where lower variance in motion vectors indicates smoother transitions.

\subsection{Mask Design}
\begin{figure}[h]
  \centering
    \includegraphics[width=1\linewidth]{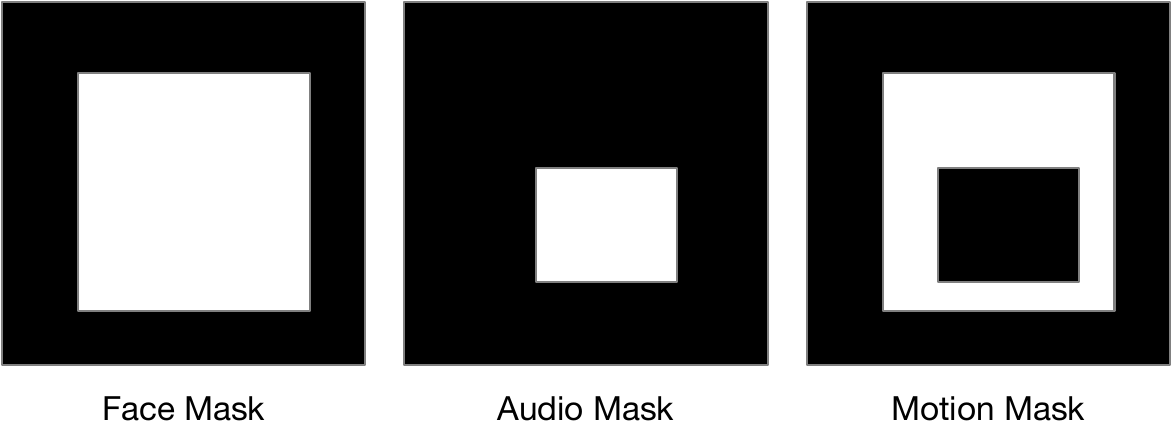}
    \caption{The type of masks we used in our framework.
    }
    \label{fig:mask}    
\end{figure}

\begin{figure*}[htbp]
  \centering
    \includegraphics[width=1\linewidth]{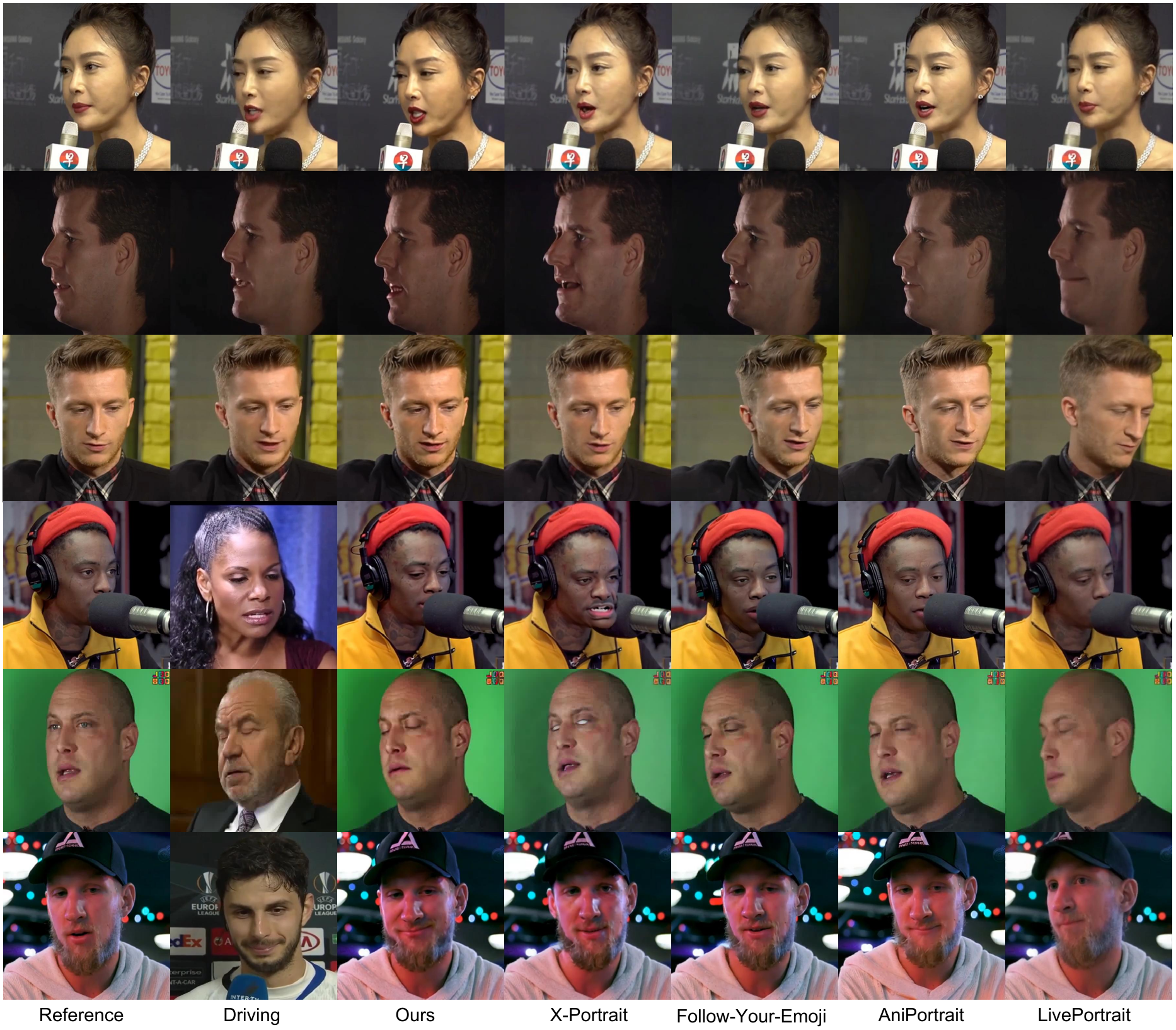}
    \caption{
    The results generated by our method under facial motion control.
    }
    \label{fig:face-compare-2}    
\end{figure*}

In our framework, we utilize masks to indicate the control regions of each signal. Three types of masks are used in our framework. As illustrated in Figure~\ref{fig:mask}, the face mask is used to indicate the rough position of the face in the source image. During training, we use RetinaFace~\cite{serengil2020lightface} to calculate the bounding box for all frames in the ground truth segments and obtain the smallest enclosing rectangle of these bounding boxes. We then draw the face mask based on that rectangle to indicate the facial location in the desired video. Similarly, the audio mask is obtained by detecting the mouth bounding boxes, and the motion mask is generated by using the face mask to minimize the audio mask. During the inference stage, we detect the bounding box of the source image and apply the appropriate extension.

\begin{figure*}[htbp]
  \centering
    \includegraphics[width=1\linewidth]{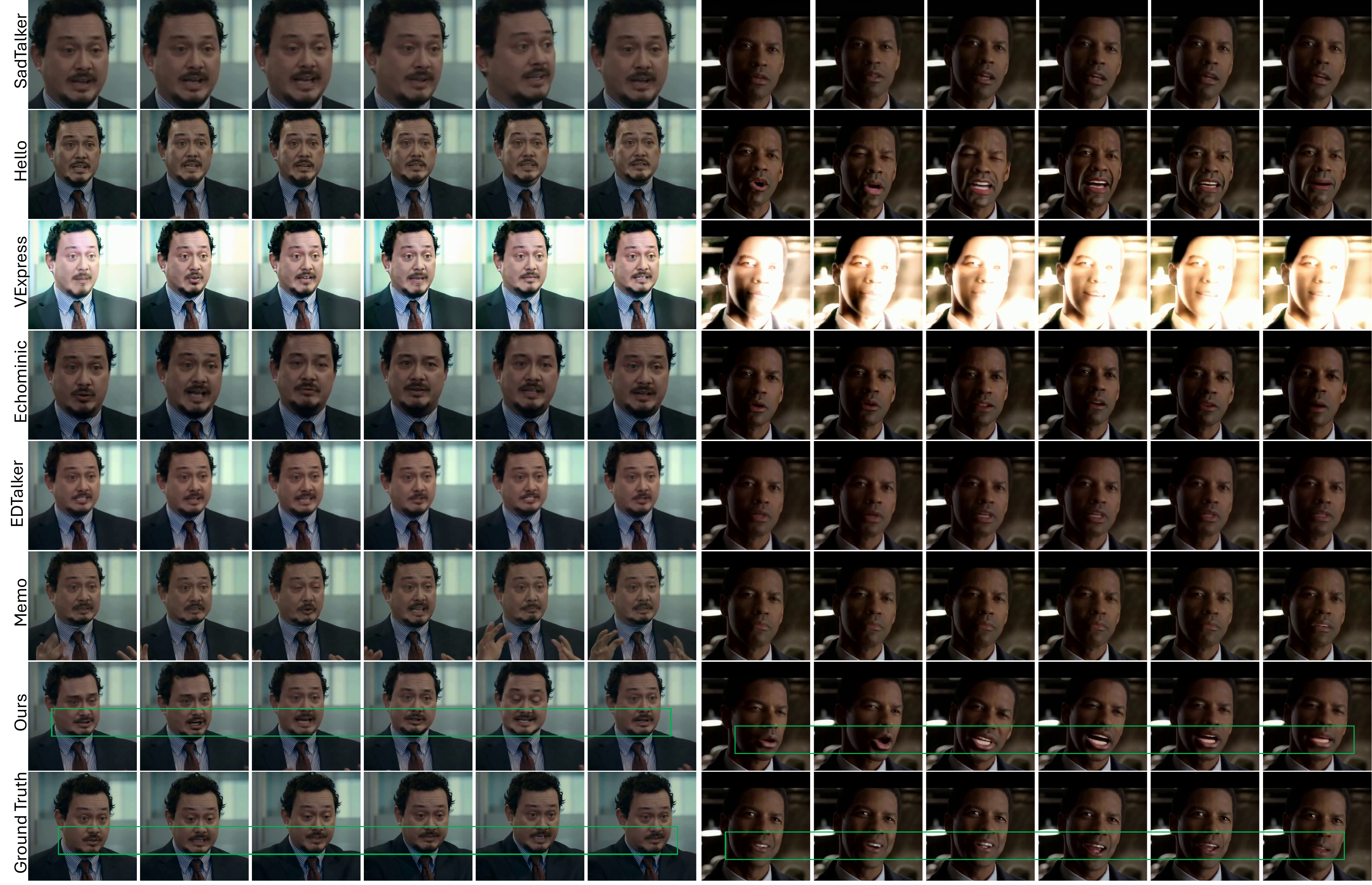}
    \caption{
    The results generated by our method under audio control.
    }
    \label{fig:supp-audio}    
\end{figure*}
\begin{figure*}[htbp]
  \centering
    \includegraphics[width=1\linewidth]{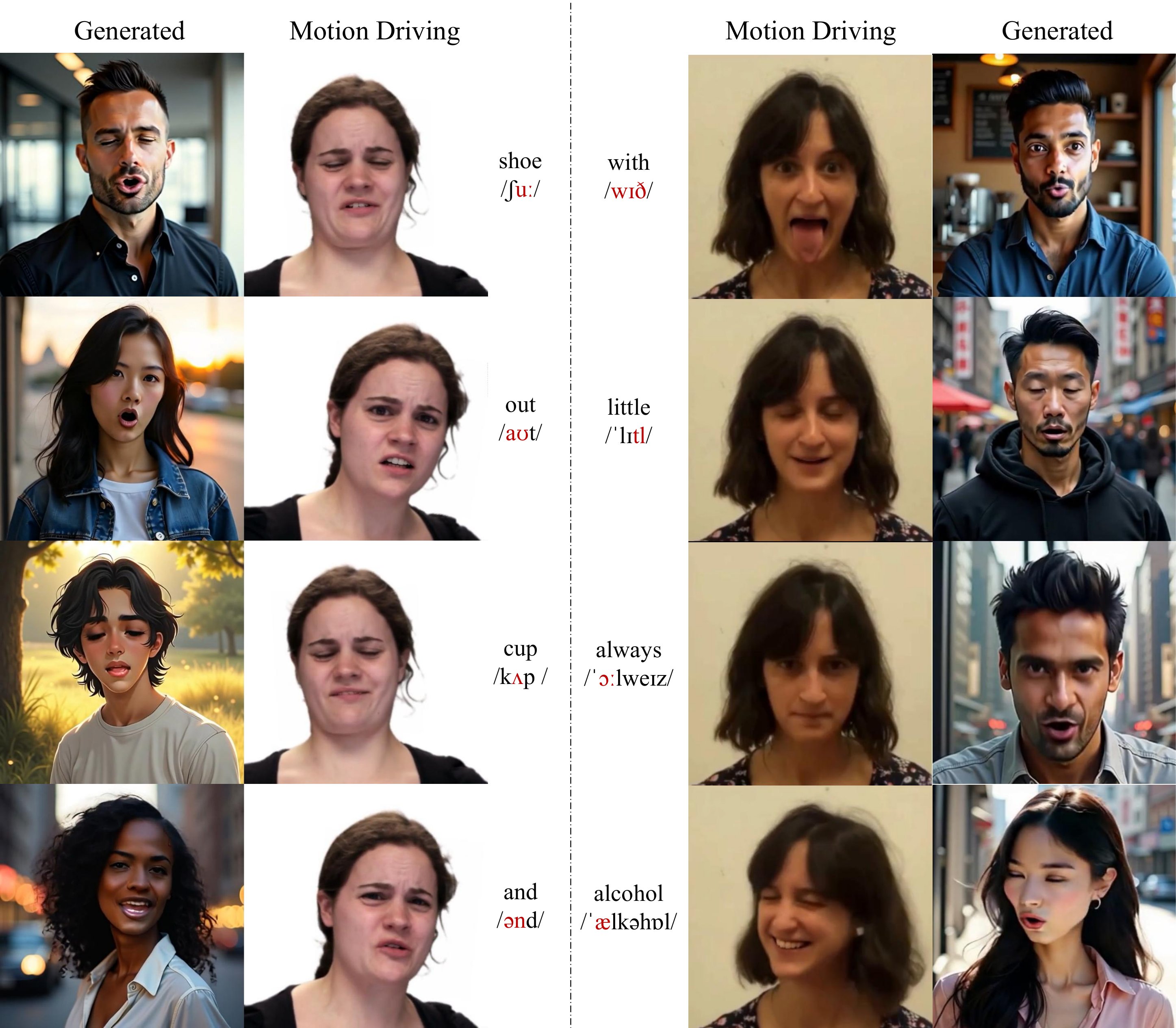}
    \caption{
    The results generated by our method under audio-visual joint control.
    }
    \label{fig:supp-1}    
\end{figure*}

\section{Visualization}
\subsection{Face reenacment}
Figure~\ref{fig:face-compare-2} demonstrates that our approach achieves enhanced precision in replicating portrait motions that align closely with the driving video's dynamics. For self-reenactment, the results generated by our framework better preserve intricate facial behaviors, particularly in eye movement patterns, ocular orientation, and lip articulation accuracy.

As illustrated in the third, fifth, and sixth rows of Figure~\ref{fig:face-compare-2}, our method can achieve tracking of the overall rotation of the head, which cannot be achieved by previous top-performing warping-based methods, such as LivePortrait.

While previous diffusion-based methods demonstrate notable advantages in output fidelity, their reliance on facial keypoint tracking introduces limitations. As shown in the third and fifth rows of Figure~\ref{fig:face-compare-2}, discrepancies in facial geometry between source and target identities, combined with the inherent limitations of keypoint representations in capturing detailed facial expressions, make previous state-of-the-art methods (e.g., AniPortrait~\cite{wei2024aniportrait}, Follow Your Emoji~\cite{ma2024follow})less effective than our method in reconstructing facial contours, gaze direction, and lip synchronization accuracy.
These keypoint-dependent methodologies remain susceptible to interference from driving video subjects' facial geometries, resulting in incomplete motion-identity separation. 
These methods face challenges in identity preservation due to changes in facial geometry resulting from misalignment of key points.
Our framework overcomes these limitations through a parallel-control mamba layer (PCM), with an improved separation of facial identity characteristics from motion parameters, as evidenced in Figure~\ref{fig:face-compare-2}. This enhanced decoupling enables superior identity retention while capturing nuanced facial dynamics.
Although X-Portrait~\cite{xie2024x} utilizes a non-explicit keypoint control method, it does not completely decouple motion and appearance information. This limitation results in noticeable flaws in the generated results, particularly evident in the fourth and fifth lines of Figure~\ref{fig:face-compare-2}.

Moreover, frameworks built upon Stable Diffusion's image generation architecture typically under-perform our method in temporal coherence metrics. By integrating the stable video diffusion model with our framework, we achieve significant improvements in three critical aspects: identity consistency preservation, visual quality optimization, and micro-expression reproduction. This collectively produces more natural-looking and temporally stable animations.
We provide video demos in the supplementary materials. In these video demos, we compare our method with other methods, and our method obviously achieves better results. Additionally, we found that when some of the reference images provide more details, the results can be even more realistic. 

\subsection{Audio Driven Talker Head Generation}
We present a comprehensive comparison with all baseline methods in Figure~\ref{fig:supp-audio}. As shown in the figure, our method is able to produce accurate lip motion while containing fewer artifacts. Notably, our method generates natural head poses and expressions similar to the ground truth (please refer to the video demo in \emph{Supplementary Material}), whereas other methods mainly manipulate the mouth shape and leave other regions static. These results confirm that our mamba design effectively aggregates audio signals with facial tokens to produce natural expressions and accurate lip synchronization, as we use the face mask as an audio mask to incorporate nearly all facial tokens in an audio-driven manner.

\subsection{Audio-visual Joint Driven}
We also present additional demonstrations in Figure~\ref{fig:supp-1}, which displays the results produced by our method under audio-visual joint control. Our approach effectively maintains lip synchronization with the audio while accurately reflecting the expressions of the Motion Driving sources.
We highly recommend watching the video demonstrations. Additional results can be found in the \emph{supplementary materials}, where video demonstrations are also available.

\section{Ethics Considerations and AI Responsibility}

This study aims to develop artificial intelligence-driven virtual avatars with enhanced visual emotional expression capabilities, utilizing audio or visual inputs, for applications in positive and constructive domains. The technology is designed specifically for ethical purposes, focusing on applications that are beneficial to society, and is not intended for generating deceptive or harmful media content.

However, as with all generative approaches in this field, there remains a theoretical concern about potential misuse for identity replication or malicious purposes. The research team strongly condemns any attempts to use the technology for creating fraudulent, harmful, or misleading representations of real individuals. Rigorous technical evaluations of the current system indicate that the generated outputs exhibit clear artificial features, and quantitative comparisons with genuine human recordings show measurable discrepancies, ensuring that the results remain distinguishable from authentic human expressions.




\end{document}